\documentclass[lettersize,journal]{IEEEtran}
\usepackage{amsmath,amsfonts}
\usepackage{booktabs}
\usepackage{subcaption}
\usepackage{textcomp}
\usepackage{stfloats}
\usepackage{url}
\usepackage{verbatim}
\usepackage{graphicx}
\usepackage{multirow}
\usepackage{tabularx}
\usepackage[table]{xcolor}
\definecolor{lightgray}{gray}{0.9}
\usepackage{xcolor}

\usepackage{arydshln}
\usepackage{amssymb}
\def\BibTeX{{\rm B\kern-.05em{\sc i\kern-.025em b}\kern-.08em
    T\kern-.1667em\lower.7ex\hbox{E}\kern-.125emX}}
\usepackage{balance}

\begin{document}
\title{View Invariant Learning for Vision-Language Navigation in Continuous Environments}
\author{Josh Qixuan Sun, Huaiyuan Weng, Xiaoying Xing, Chul Min Yeum, and Mark Crowley
\thanks{Josh Qixuan Sun and Mark Crowley are with the Department of Electrical and Computer Engineering, University of Waterloo, Waterloo N2L 3G1, Canada. Huaiyuan Weng and Chul Min Yeum are with the Department of Civil and Environmental Engineering, University of Waterloo, Waterloo N2L 3G1, Canada. Emails: josh.q.sun, mark.crowley, huaiyuan.weng, cmyeum@uwaterloo.ca. Xiaoying Xing (xiaoyingxing2026@u.northwestern.edu) is with the Department of Electrical and Computer Engineering, Northwestern University, Evanston, IL 60208 USA. $^*$Corresponding Author: Josh Qixuan Sun.}}

\markboth{Journal of \LaTeX\ Class Files,~Vol.~18, No.~9, September~2020}%
{How to Use the IEEEtran \LaTeX \ Templates}

\maketitle

\thispagestyle{empty}
\pagestyle{empty}
\markboth{}{}

\begin{abstract}
Vision-Language Navigation in Continuous Environments (VLNCE), where an agent follows instructions and moves freely to reach a destination, is a key research problem in embodied AI. However, most existing approaches are sensitive to viewpoint changes, i.e. variations in camera height and viewing angle. Here we introduce a more general scenario, V$^2$-VLNCE (VLNCE with Varied Viewpoints) and propose a view-invariant post-training framework, called VIL (View Invariant Learning), that makes existing navigation policies more robust to changes in camera viewpoint. VIL employs a contrastive learning framework to learn sparse and view-invariant features. We also introduce a teacher-student framework for the Waypoint Predictor Module, a standard part of VLNCE baselines, where a view-dependent teacher model distills knowledge into a view-invariant student model. We employ an end-to-end training paradigm to jointly optimize these components. Empirical results show that our method outperforms state-of-the-art approaches on V$^2$-VLNCE by 8-15\% measured on Success Rate for two standard benchmark datasets R2R-CE and RxR-CE. 
Evaluation of VIL in standard VLNCE settings shows that despite being trained for varied viewpoints, VIL often still improves performance. 
On the harder RxR-CE dataset, our method also achieved state-of-the-art performance across all metrics. 
This suggests that adding VIL does not diminish the standard viewpoint performance and can serve as a plug-and-play post-training method. 
We further evaluate VIL for simulated camera placements derived from real robot configurations (e.g. Stretch RE-1, LoCoBot), showing consistent improvements of performance. Finally, we present a proof-of-concept real-robot evaluation in two physical environments using a panoramic RGB sensor combined with LiDAR. These results show that VIL improves robustness not only in simulation but also in real-world navigation scenarios, making it a practical strategy for embodied agents. The code is available at https://github.com/realjoshqsun/V2-VLNCE.
\end{abstract}


\section{Introduction}

 \textit{Vision-Language Navigation (VLN)} \cite{anderson2018vision} requires an agent to follow human instructions by executing a sequence of actions. Traditional VLN assumes a predefined topological graph, where the agent moves along predefined graph edges. The broader \textit{VLN in Continuous Environments (VLNCE)} \cite{krantz2020beyond-vlnce1} problem removes this constraint, enabling agents to move freely in continuous space. Prior works in VLNCE mainly improve navigation by designing better neural architectures \cite{hong2021vln-vlnbert,georgakis2022cross-cm2,wang2023gridmm,an2024etpnav} or incorporating richer spatial and semantic features \cite{hong2022bridging,wang2024lookahead-vlnnerf,wang2024sim,zhang2024narrowing}.
However, these learned policies often struggle when the camera viewpoint (i.e. height and viewing angle) changes during deployment where even small shifts in viewpoint can lead to large performance drops. 

\begin{figure}[t]
    \centering
    \includegraphics[width=\linewidth]{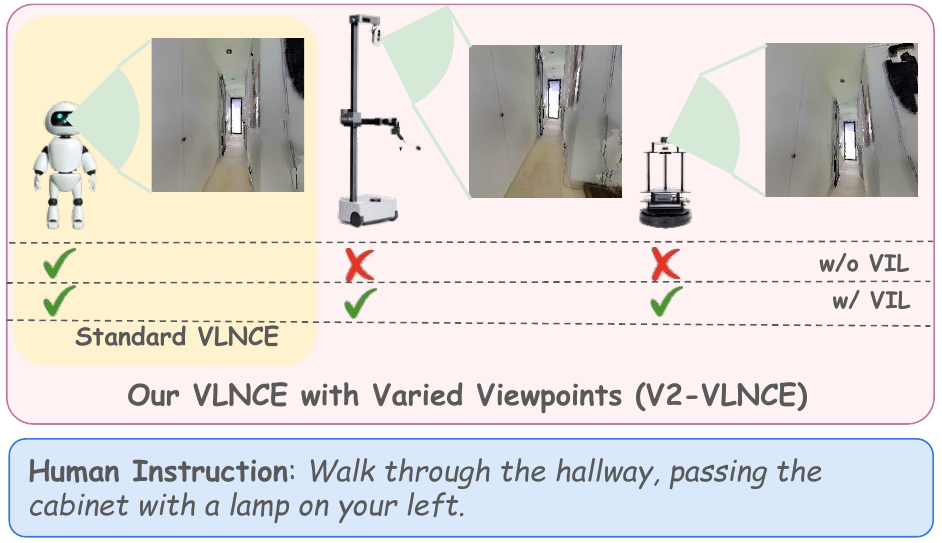}
    \caption{{Comparison between standard VLNCE and our proposed V$^2$-VLNCE. Under viewpoint changes, baseline navigation policies suffer from degraded performance. Applying View Invariant Learning (VIL) significantly improves robustness, enabling the agent to navigate under varied viewpoints.}}
    \label{fig:fig1}
\end{figure}

This \textit{Varied Viewpoint} challenge, occurs when agents need to generalize across environments with different egocentric camera placements and is especially relevant in real-world robotics, where robots have different camera mounting positions. To systematically study this problem, we introduce \textit{V$^2$-VLNCE (VLNCE with Varied Viewpoints)}, a generalized setting designed to evaluate policy performance under diverse camera viewpoints. We focus on two variables: camera height and angle. In each episode, we sample a viewpoint from a 2D distribution over a range of heights and angles which better reflect the variation in real-world scenarios. While prior work \cite{seo2023multi,liu2024robouniview} considered this varied viewpoint challenge, they were developed for robotic manipulation rather than navigation tasks. As for VLNCE, GVNav \cite{li2025ground} focused on the impact of height shift and addressed by training with this specific configuration. However, that approach only considers a single, fixed camera height and does not account for variations in both heights and angles simultaneously as we do. 

While prior work has touched on aspects of viewpoint variation, their solutions are either not applicable to navigation or suffer from significant inefficiencies. For robotic manipulation, methods like MV-MVM \cite{seo2023multi}, RoboUniView \cite{liu2024robouniview}, and ReViWo \cite{panglearning} address varied viewpoints, but only for manipulation tasks. They also often rely on extensive pre-training to learn robust representations, which are then fine-tuned for specific downstream tasks. Thus, this paradigm is computationally intensive and not easily transferable. For VLNCE, GVNav \cite{li2025ground} focuses on the impact of viewpoint changes by adopting a single, fixed ground-level viewpoint and training their model from scratch with this specific configuration. Thus, they cannot account for more complex, continuous variations in both heights and angles. As we will demonstrate in Section \ref{sec:ablation}, this simple retraining strategy proves insufficient to handle the more demanding V$^2$-VLNCE setting. These limitations show a critical need for more computationally efficient and generalizable approaches. Instead of costly retraining for each new viewpoint, we develop a single policy that can be adapted to diverse viewpoints with minimal effort.

We propose \textit{View Invariant Learning (VIL)}, a strategy that adapts existing policies to varied viewpoints without retraining from scratch. VIL consists of two components: a contrastive learning objective and a teacher-student framework for waypoint prediction. The contrastive framework encourages the policy to learn sparse, view-invariant features by aligning representations from different viewpoints of the same scene, while separating unrelated observations. Features used for contrastive learning are extracted through a projection head, and the learned representations are shared with the navigation policy. For waypoint prediction, a frozen teacher model, initialized from a pretrained policy, processes observations from a standard viewpoint. The student model shares the same architecture as the teacher but trains only a small adapter module inserted into the waypoint predictor, while freezing the rest of the weights. The student, receiving varied-viewpoint inputs, learns to match the teacher’s outputs through a distillation loss. Both components are trained jointly and end-to-end to enable efficient viewpoint adaptation.

Our contributions are as follows:
\textbf{1)} We introduce V$^2$-VLNCE, a new evaluation setting that incorporates both camera height and viewing angle variations to simulate diverse camera viewpoints. This setting enables a more realistic and systematic analysis of viewpoint robustness.
\textbf{2)} We propose VIL, a strategy trained with diverse viewpoints using a contrastive learning objective and a teacher-student framework.
\textbf{3)} We conduct extensive experiments in simulation, showing that VIL outperforms existing baselines in the V$^2$-VLNCE setting.
\textbf{4)} We further evaluate VIL under simulated camera placements derived from real robot configurations, confirming robustness across practical embodiment settings.

\section{Related Work}

\textbf{Vision-language navigation.} Prior studies on VLNCE \cite{krantz2020beyond-vlnce1} have focused on enhancing input modalities through a variety of methods, including panoramic RGB-D images \cite{hong2021vln-vlnbert,wang2023gridmm,an2024etpnav,hong2022bridging,wang2024lookahead-vlnnerf}, semantic information \cite{georgakis2022cross-cm2,wang2024sim,hong2023learning,irshad2022semantically}, occupancy maps \cite{zhang2024narrowing,safevln}, and larger-scale training data \cite{wang2023scaling,wang2025bootstrapping}. Other works focus on designing more efficient neural networks for vision-and-language fusion \cite{wang2023dreamwalker,raychaudhuri2021language,zhang2024navid,zhang2024uni,zhang2025embodied}. 

Waypoint predictors are crucial to recent VLNCE models \cite{an2024etpnav,hong2022bridging,wang2025bootstrapping}, as they bridge VLN and VLNCE and enable pre-training on VLN. To adapt to ground-level views, GVNav \cite{li2025ground} retrained the waypoint predictor separately with matching data. In contrast, we train the entire model end-to-end, removing the need for separate waypoint predictor training.

\textbf{Varied viewpoint challenge in robotics.} In robotic manipulation, several works focus on learning view-invariant representations to address viewpoint variation \cite{seo2023multi, liu2024robouniview, panglearning}. However, these approaches usually adopt a two-stage training pipeline: first learning a view-invariant encoder, then training the policy on top of the frozen encoder. This strategy is less suitable for VLNCE. First, VLNCE policies are typically pretrained on VLN datasets, and applying a two-stage pipeline would discard the benefits of this pretraining. Second, the training cost would be high. Our goal is not to retrain new policies from scratch, but to adapt existing policies to varied viewpoints. Third, VLNCE architectures often include a waypoint predictor, which would also require separate training in a two-stage pipeline, further increasing the cost.

In robotic navigation, several recent works have explored viewpoint robustness under different task settings.  GVNav \cite{li2025ground} studies ground-level viewpoint variation for VLNCE by retraining both the navigation policy and the waypoint predictor on a fixed low-height camera configuration.  In contrast, our work considers a more general viewpoint setting by modeling a joint distribution over camera height and pitch angle, which better reflects realistic camera mounting variations. 
Moreover, GVNav relies on a decoupled training scheme with separate optimization of the waypoint predictor and the policy, whereas our VIL framework enables efficient end-to-end adaptation through lightweight adapters without retraining the core pretrained policy. 
RING \cite{eftekhar2024one}, a concurrent work, investigates viewpoint robustness for the ObjectNav benchmark by randomizing camera configurations during training. 
However, ObjectNav aims a simple semantic label (e.g., "find a mug"), while VLNCE requires the agent to faithfully follow long-form, multi-step linguistic instructions and reach specific subgoals along a trajectory and demands much finer alignment between language and visual perspective. 
Methodologically, RING follows a domain randomization strategy, whereas our approach introduces two architecture-compatible modules, contrastive learning and waypoint predictor distillation, to explicitly enforce viewpoint invariance while preserving pretrained VLNCE knowledge.

\section{Method}

\begin{figure*}[t]
    \centering
    \includegraphics[width=1.0\linewidth]{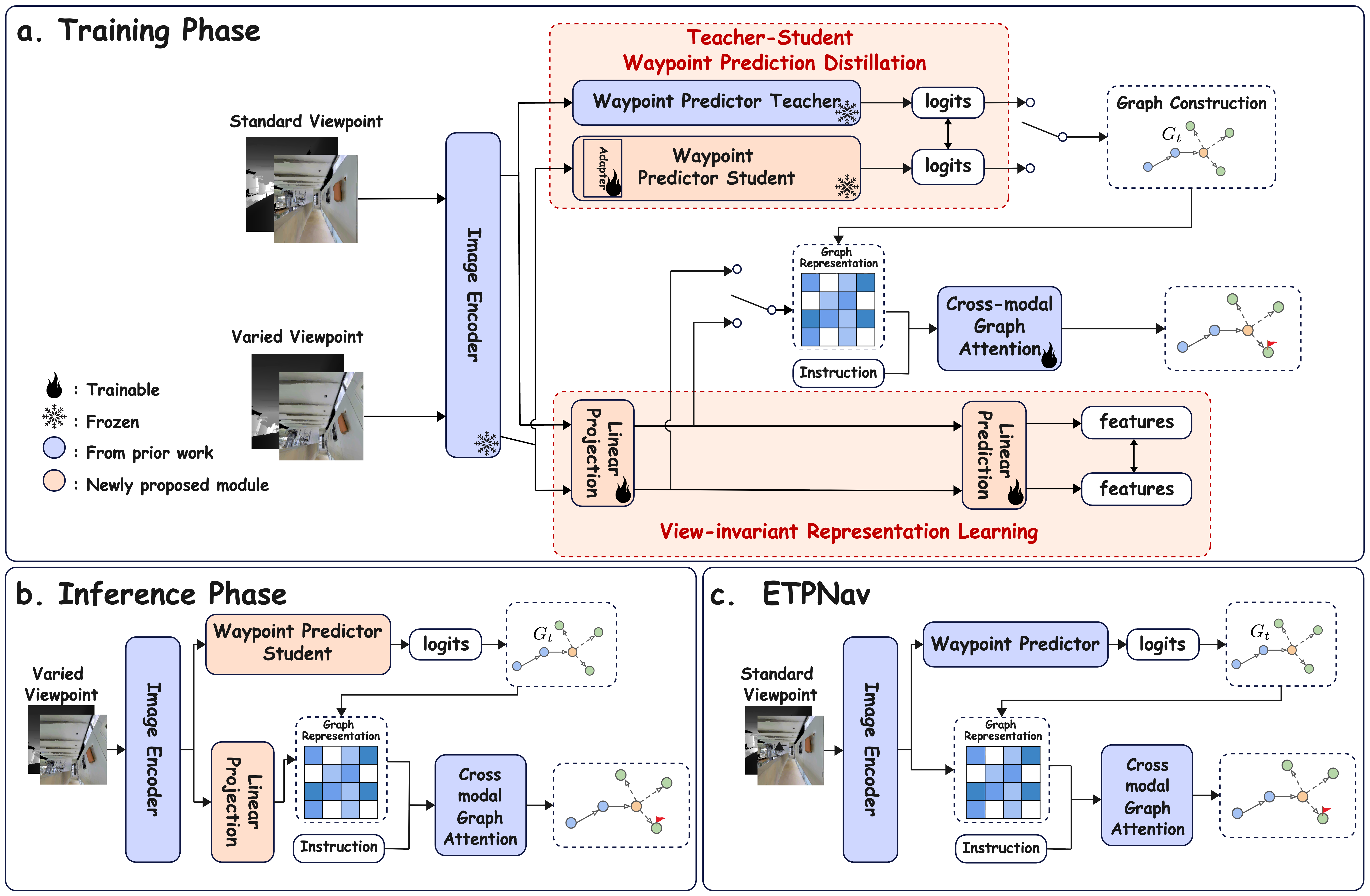}
    \caption{
        Overview of our view-invariant learning framework. 
        \textbf{(a) Training Phase}: Given standard and varied viewpoints, the image encoder extracts features for both. A contrastive learning objective is applied to align representations across viewpoints and encourage view-invariant features. 
        Meanwhile, a teacher-student framework is used for waypoint prediction, where a frozen teacher processes standard views and a student model adapts to varied views by training only a lightweight adapter module.
        \textbf{(b) Inference Phase}: Only the student model is used to predict waypoints under varied viewpoints. 
        \textbf{(c) ETPNav baseline}: A standard VLNCE architecture without contrastive learning or teacher-student training.
    }
    \label{fig:method}
\end{figure*}

\subsection{ETPNav Preliminary}\label{sec:etpnav_pre}
We build on ETPNav \cite{an2024etpnav}, a strong panoramic VLNCE baseline. At each step $t$, the agent receives a natural language instruction and panoramic RGB-D observations $O_t = \{O_t^{\text{rgb}}, O_t^{\text{d}}\}$ consisting of 12 RGB and 12 depth views captured at equally spaced $30^\circ$ intervals. As illustrated in Fig.~\ref{fig:method}(c), ETPNav predicts navigable waypoint candidates from the panoramic inputs, incrementally builds a local topological map, fuses instruction embeddings with graph representations through cross-modal graph attention, and finally selects the next navigation target. For details of the full architecture, we refer readers to \cite{an2024etpnav}.

\subsection{View-invariant Representation Learning}\label{sec:contrastive}

Policies trained under a fixed camera configuration often fail to generalize when the viewpoint changes. To address this, we introduce a contrastive learning objective that encourages learning of viewpoint-invariant features and integrates directly into the navigation model.

Given a panoramic RGB-D observation $O_t$ at time step $t$, we generate two views of the scene: a standard viewpoint $O_t^{\text{std}}$ and a varied viewpoint $O_t^{\text{var}}$, created by randomly shifting the camera height and angle. Both views are encoded by a shared visual encoder $f_{\text{enc}}(\cdot)$.

We apply a three-layer projection head $g(\cdot)$ after the encoder, following the standard design of SimCLRv2 \cite{chen2020big}. We denote the output of the first linear layer as $g_1(\cdot)$, the second as $g_2(\cdot)$, and the third as $g_3(\cdot)$. The features used for downstream navigation and contrastive learning are:
\[
f_{\text{task}} = g_1(f_{\text{enc}}(O_t)), \quad f_{\text{contrast}} = g_3(g_2(g_1(f_{\text{enc}}(O_t))))
\]
We further distinguish the task features from different viewpoints. Let $f_{\text{task}}^{\text{std}}$ denote the task feature from the standard viewpoint and $f_{\text{task}}^{\text{var}}$ denote that from the varied viewpoint. We construct a graph representation by sampling either $f_{\text{task}}^{\text{std}}$ or $f_{\text{task}}^{\text{var}}$ with probability $p_1$, and combine the selected features with the topological graph $G_t$ to represent the current scene structure. For contrastive learning, we also denote the features from the two viewpoints separately as $f_{\text{contrast}}^{\text{std}}$ and $f_{\text{contrast}}^{\text{var}}$, which are used to compute the contrastive loss across viewpoints.

To ensure compatibility with the pretrained ETPNav model, we initialize the first linear layer $g_1$ as an identity matrix. This initialization preserves the original feature distribution at the beginning of training and allows gradual adaptation to varied viewpoints.


The contrastive learning objective enforces feature consistency between the standard and varied views of the same scene. For each instance in a training batch, indexed by $(i, j)$ where $i$ denotes the batch index and $j$ denotes the panoramic view index, where $j \in [0, 1, \ldots, 11]$, corresponding to $[0^\circ, 30^\circ, \ldots, 330^\circ]$ heading angles., we define positive pairs as the features corresponding to the same heading $j$ under standard and varied viewpoints. Negative pairs are constructed from two sources: (1) random cross-scene negatives sampled from different scenes. For implementation efficiency, the latter is achieved by shifting indices within the mini-batch, i.e., $((i+1) \pmod{\text{batch\_size}}, j)$, and (2) intra-scene hard negatives by selecting features from the opposite heading $(i, (j+6) \pmod{12})$ The contrastive loss follows the standard InfoNCE formulation \cite{chen2020big}:
\[
\mathcal{L}_{\text{cl}} = -\log \frac{\exp(\text{sim}(q, k^+)/\tau)}{\exp(\text{sim}(q, k^+)/\tau) + \sum_{k^-} \exp(\text{sim}(q, k^-)/\tau)},
\]
where $q$ is the feature of the standard view, $k^+$ is the feature of the varied view of the same scene, $k^-$ are features from negative samples, and $\text{sim}(\cdot, \cdot)$ denotes cosine similarity. $\tau$ is the temperature parameter, set to $1.0$ following standard practice.  By jointly optimizing this contrastive objective with the navigation policy, the agent learns feature representations that are more robust to viewpoint variations without sacrificing performance on the original downstream task.

\subsection{Teacher-Student Waypoint Prediction Distillation}\label{sec:teacher-student}

\begin{figure}[t]
    \centering
    \includegraphics[width=\linewidth]{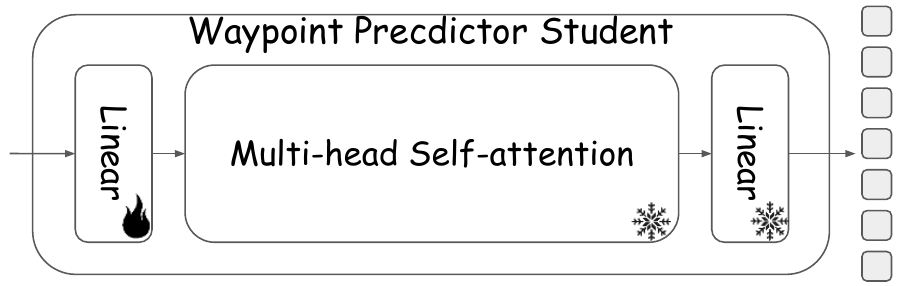}
    \caption{Detailed architecture of the waypoint predictor student module used in teacher–student distillation.}
    \label{fig:wpd}
\end{figure}

In VLN tasks, the quality of waypoint prediction is critical for navigation success. Recent works such as GVNav \cite{li2025ground} have observed that waypoint predictors trained under the standard viewpoint experience significant performance degradation when evaluated from a ground-level viewpoint. GVNav addresses this issue by retraining the waypoint predictor separately with ground-level viewpoint data, but this two-stage training strategy incurs a high training cost. In contrast, we propose an integrated teacher-student framework that enables robust waypoint prediction under varied viewpoints without additional training stages.

The teacher and student models share the same waypoint predictor architecture, introduced in Section \ref{sec:etpnav_pre}, where a transformer-based network predicts a dense heatmap of nearby navigable waypoints from panoramic RGB-D observations. Both the teacher and student models are initialized from the one used in ETPNav \cite{an2024etpnav}. As shown in Figure \ref{fig:method}(a), the teacher model is frozen, and operates on standard viewpoint observations. The student model processes varied viewpoint observations and adapts via lightweight adapter layers, while the rest of the weights are frozen. 
Figure~\ref{fig:wpd} illustrates the detailed architecture of the waypoint predictor student in Figure ~\ref{fig:method}(a). 
Importantly, the adapter is implemented as the original input linear layer of the predictor, which is made trainable during distillation, rather than being inserted as an additional module into the backbone. Formally, given an observation at time $t$, the teacher outputs waypoint logits $S_t^{\text{teacher}}$, and the student outputs $S_t^{\text{student}}$. To align the student with the teacher, we apply KL divergence as the distillation loss:
\[
\mathcal{L}_{\text{wpd}} = \text{KL}\left(\text{softmax}(S_t^{\text{teacher}}) \parallel \text{softmax}(S_t^{\text{student}})\right).
\]
where both logits are normalized by softmax before computing divergence. During graph construction, we sample either the teacher or student predictions, with probability $p_2$ or $1-p_2$ respectively, to build the local topological map $G_t$.

\subsection{Training Objective}

Our full model is trained end-to-end by jointly optimizing three objectives: the standard navigation loss $\mathcal{L}_{\text{nav}}$, the contrastive learning loss $\mathcal{L}_{\text{cl}}$ introduced in Section \ref{sec:contrastive}, and the waypoint predictor distillation loss $\mathcal{L}_{\text{wpd}}$ introduced in Section \ref{sec:teacher-student}. The overall training loss is formulated as:
\[
\mathcal{L} = \mathcal{L}_{\text{nav}} + \lambda_{1} \mathcal{L}_{\text{cl}} + \lambda_{2} \mathcal{L}_{\text{wpd}}.
\]
Here, \(\lambda_1\) and \(\lambda_2\) are hyperparameters that balance the contributions of different losses.

\section{Experiments}

We aim to answer the following four research questions.  \textbf{Q1}: How does our VIL strategy perform compared to existing baseline methods under varied viewpoints? (Sec. \ref{sec:main_results}) \textbf{Q2}: Does VIL maintain performance on the original VLNCE setting? (Sec. \ref{sec:vlnce}) \textbf{Q3}: Is standard fine-tuning (i.e., retraining the model under varied viewpoint data) sufficient? What is the contribution of each component? (Sec. \ref{sec:ablation}) \textbf{Q4}: Does VIL extrapolate to out-of-distribution viewpoints? (Sec.~\ref{sec:ood}) 

\textbf{Baselines.} We evaluate evaluate our VIL strategy by applying it to two strong VLNCE baselines: BEVBert \cite{an2022bevbert} and ETPNav \cite{an2024etpnav}. Both methods demonstrate strong performance on standard VLNCE benchmarks and provide trained checkpoints, making them widely used in recent studies. We apply VIL on top of each baseline to evaluate its compatibility and performance gain across different architectures.

\textbf{Benchmarks.} The R2R-CE dataset \cite{krantz2020beyond-vlnce1} comprises a total of 5,611 trajectories and the average path length is 9.89m and each instruction is comprised of an average of 32 words. Compared to R2R-CE, RxR-CE \cite{ku2020room} is larger and more challenging. RxR-CE presents substantively longer instructions, an average of 120 words per instruction, and annotated paths in RxR-CE are much longer than those in R2R-CE with an average length of 15.32m. To evaluate generalization, the val-seen and val-unseen splits are commonly used. Both splits include navigation instructions not seen during training. The main difference lies in the environments: val-seen environments appear in the training set, while val-unseen does not.

\textbf{Metrics.} We adopt the following navigation metrics from previous works. Navigation Error (NE): average geometric distance in meters between the final and target location; Success Rate (SR): the ratio of paths with NE less than 3 meters; Oracle SR (OSR) \cite{anderson2018vision}: SR given an oracle stop policy; SR penalized by Path Length (SPL) \cite{anderson2018evaluation}; Normalized Dynamic Time Wrapping (nDTW) \cite{ilharco2019general}: a normalized DTW score between predicted and ground-truth paths; Success weighted normalized Dynamic Time Warping  (SDTW) \cite{ilharco2019general}: nDTW weighted by success. 

\textbf{Implementation details.} 
All models are initialized from the official pretrained checkpoints of the corresponding backbones, and all hyperparameters of the backbone models follow their original implementations, while only a small set of VIL-specific hyperparameters is introduced. The hyperparameters specific to VIL are: sampling probabilities $p_1 = p_2 = 0.1$, contrastive learning loss weight $\lambda_1 = 0.2$, and waypoint predictor distillation loss weight $\lambda_2 = 10.0$.

\subsection{Performance under varied viewpoints}\label{sec:main_results}

\begin{table*}[t]
\centering
\caption{
Comparison on R2R-CE and RxR-CE under the \textit{Varied Viewpoint} and \textit{Ground-level Viewpoint} settings. The \textit{Varied Viewpoint} setting corresponds to our proposed V$^2$-VLNCE setup. The \textit{Ground-level Viewpoint} setting is adapted from GVNav \cite{li2025ground}. Evaluation metrics are consistent with GVNav. \textbf{Bold} indicates performance improvements introduced by VIL.
}
\begin{tabular*}{0.8\textwidth}{@{\extracolsep{\fill}}lcccccccccc}
\toprule
\multirow{2}{*}{Method} & \multicolumn{5}{c}{val-seen} & \multicolumn{5}{c}{val-unseen} \\
\cmidrule(lr){2-6} \cmidrule(lr){7-11}
 & NE$\downarrow$ & nDTW$\uparrow$ & OSR$\uparrow$ & SR$\uparrow$ & SPL$\uparrow$ & NE$\downarrow$ & nDTW$\uparrow$ & OSR$\uparrow$ & SR$\uparrow$ & SPL$\uparrow$  \\
\midrule
\rowcolor{lightgray}
\multicolumn{11}{l}{R2R-CE, \textit{Ground-level Viewpoint} \cite{li2025ground}} \\
HPN \cite{krantz2021waypoint}   \textcolor{gray}{\scalebox{0.7}{[ICCV2021]}} & 6.30 & 57 & 43 & 37 & 35 & 6.79 & 54 & 35 & 30 & 28 \\
CMA \cite{hong2022bridging} \textcolor{gray}{\scalebox{0.7}{[CVPR2022]}} & 5.99 & 55 & 58 & 44 & 38 & 6.68 & 49 & 50 & 37 & 30 \\
VLN$\circlearrowright$BERT \cite{hong2022bridging} \textcolor{gray}{\scalebox{0.7}{[CVPR2022]}} & 5.46 & 55 & 56 & 47 & 39 & 6.25 & 50 & 49 & 39 & 33 \\
GVNav \cite{li2025ground} \textcolor{gray}{\scalebox{0.7}{[ICRA2025]}}  & 3.88 &66 &70 &64 &56 & 4.89 &58 &62 &55 &45\\
\hline
BEVBert \cite{an2022bevbert} \textcolor{gray}{\scalebox{0.7}{[ICCV2023]}} & 3.26 & 70 & 76 & 70 & 62 & 4.63 & 61 & 67 & 59 & 49 \\
BEVBert + VIL (Ours) & \textbf{3.16} & \textbf{71} & \textbf{77} & \textbf{71} & \textbf{63} & \textbf{4.61} & \textbf{62} & 66 & 59 & \textbf{50} \\
\hdashline
ETPNav \cite{an2024etpnav} \textcolor{gray}{\scalebox{0.7}{[TPAMI2024]}} & 4.48 & 62 & 71 & 62 & 50 & 5.27 & 55 & 63 & 52 & 42\\
ETPNav + VIL (Ours) & \textbf{4.02} & \textbf{67} & \textbf{71} & \textbf{64} & \textbf{55} &
\textbf{4.91} & \textbf{59} & \textbf{65} & \textbf{57} & \textbf{47} \\
\midrule
\rowcolor{lightgray}
\multicolumn{11}{l}{R2R-CE, \textit{Varied Viewpoint (Ours)}} \\
HPN \cite{krantz2021waypoint}   \textcolor{gray}{\scalebox{0.7}{[ICCV2021]}} & 6.32 & 57 & 43 & 35 & 33 & 6.76 & 54 & 35 & 29 & 27 \\
CMA \cite{hong2022bridging} \textcolor{gray}{\scalebox{0.7}{[CVPR2022]}} & 6.59 & 49 & 45 & 32 & 27 & 6.91 & 46 & 40 & 28 & 23 \\
VLN\(\circlearrowright\)BERT \cite{hong2022bridging} \textcolor{gray}{\scalebox{0.7}{[CVPR2022]}} & 5.93 & 52 & 50 & 39 & 34 & 6.39 & 48 & 44 & 32 & 27 \\
VLN-3DFF \cite{wang2024sim} \textcolor{gray}{\scalebox{0.7}{[CoRL2024]}} & 5.59 & 49 & 54 & 42 & 32 & 6.12 & 45 & 54 & 41 & 31 \\
g3D-LF \cite{wang2024g3d} \textcolor{gray}{\scalebox{0.7}{[CVPR2025]}} & 5.06 & 58 & 57 & 51 & 41 & 5.26 & 56 & 57 & 50 & 40 \\
\hline
BEVBert \cite{an2022bevbert} \textcolor{gray}{\scalebox{0.7}{[ICCV2023]}} & 4.48 & 61 & 65 & 57 & 47 & 5.32 & 56 & 58& 49 & 39 \\
BEVBert + VIL (Ours) & \textbf{3.91} & \textbf{67} & \textbf{70} & \textbf{63} & \textbf{55} & \textbf{5.15} & \textbf{58} & \textbf{62} & \textbf{52} & \textbf{44} \\
\hdashline
ETPNav \cite{an2024etpnav} \textcolor{gray}{\scalebox{0.7}{[TPAMI2024]}} & 5.16 & 59 & 58 & 49 & 42 & 5.58 & 55 & 55 & 47 & 38\\
ETPNav + VIL (Ours) & \textbf{4.02} & \textbf{66} & \textbf{69} & \textbf{64} & \textbf{55} & \textbf{4.90} & \textbf{59} & \textbf{61} & \textbf{55} & \textbf{45}\\
\midrule
\rowcolor{lightgray}
\multicolumn{11}{l}{RxR-CE, \textit{Varied Viewpoint (Ours)}} \\
ETPNav \cite{an2024etpnav} \textcolor{gray}{\scalebox{0.7}{[TPAMI2024]}} & 8.07 & 50 & 49 & 40 & 31 & 7.82 & 49 & 48 & 39 & 31 \\
ETPNav + VIL (Ours) & \textbf{5.99} & \textbf{63} & \textbf{62} &  \textbf{55} & \textbf{46} & \textbf{6.42} & \textbf{59} & \textbf{57} & \textbf{50} & \textbf{41} \\
\bottomrule
\end{tabular*}
\label{tab:combined_r2r_rxr_ce}
\end{table*}

The \textit{Varied Viewpoint} protocol corresponds to our proposed V$^2$-VLNCE setting. Concretely, each viewpoint is defined by a height-angle pair $(h, \theta)$ sampled from a uniform distribution $\mathcal{U}([-0.5\text{m}, 0.5\text{m}]) \times \mathcal{U}([-30^\circ, 30^\circ])$, relative to the standard VLNCE configuration. This generalized setup better reflects real-world differences and tests model robustness to viewpoint shifts. As a baseline, we include GVNav~\cite{li2025ground}, which introduces a \textit{Ground-level Viewpoint} setting by lowering the camera height to 0.8 meters. Although GVNav does not account for angles and uses a fixed-height viewpoint, it is the only prior work that explicitly investigates viewpoint shift in VLNCE. We therefore consider it a relevant setting.

\textbf{Performance on R2R-CE.} Table \ref{tab:combined_r2r_rxr_ce} shows that applying VIL substantially improves performance under the \textit{Varied Viewpoint} setting. For example, ETPNav + VIL achieves significant gains over the base ETPNav model on both val-seen and val-unseen splits. Specifically, our model improves NE by 0.68-1.14, nDTW by 3\%–7\%, OSR by 6\%–9\%, SR by 8\%–15\%, and SPL by 7\%–13\%. Similarly, compared to BEVBert, our method shows consistent improvement across all metrics. These results demonstrate the effectiveness of VIL in promoting viewpoint-robust navigation behavior. Moreover, compared to GVNav, a method specifically designed for \textit{Ground-level Viewpoint}, ETPNav + VIL still performs better on val-unseen (e.g., +3\% OSR, +2\% SR, +2\% SPL). This suggests that our method can be generalized to \textit{Ground-level Viewpoint} as well, even without training on matched \textit{Ground-level Viewpoint} samples. 

\begin{table}[t]
\centering
\caption{Performance under standard viewpoint. Metrics: NE$\downarrow$, SR$\uparrow$, SPL$\uparrow$. Bold indicates improvement from VIL.}
\label{tab:std-viewpoint}
\resizebox{\linewidth}{!}{%
\begin{tabular}{lcccccc}
\toprule
\multirow{2}{*}{Method} & \multicolumn{3}{c}{val-seen} & \multicolumn{3}{c}{val-unseen}  \\
\cmidrule(lr){2-4} \cmidrule(lr){5-7} 
 & NE$\downarrow$ & SR$\uparrow$ & SPL$\uparrow$  & NE$\downarrow$ & SR$\uparrow$ & SPL$\uparrow$ \\
 \midrule
 \rowcolor{lightgray}
\multicolumn{7}{l}{R2R-CE} \\
VLN\(\circlearrowright\)BERT \cite{hong2022bridging} \textcolor{gray}{\scalebox{0.7}{[CVPR2022]}} & 5.02 & 50 & 44 & 5.74 & 44 &39 \\
ENP \cite{liu2024vision}  \textcolor{gray}{\scalebox{0.7}{[NeurIPS2024]}} & 3.90 & 68 & 59 & 4.69 & 58 & 50 \\
NaVILA \cite{cheng2024navila} \textcolor{gray}{\scalebox{0.7}{[RSS2025]}} & -& -& -& 5.22 &54 &49 \\
\hline
BEVBert \cite{an2022bevbert} \textcolor{gray}{\scalebox{0.7}{[ICCV2023]}} & 3.24 & 70.9 & 62.8 & 4.63 & 59.1 & 49.2\\
BEVBert + VIL (Ours) & \textbf{3.16} & \textbf{71.1} & \textbf{63.0} & \textbf{4.61} & 58.6 & \textbf{49.6}\\ \hdashline
ETPNav \cite{an2024etpnav} \textcolor{gray}{\scalebox{0.7}{[TPAMI2024]}} & 3.97 & 65.8 & 59.2 & 4.78 & 56.8 & 48.9 \\
ETPNav + VIL (Ours) & \textbf{3.71} & \textbf{67.6} & \textbf{60.4} & \textbf{4.69} & \textbf{58.3} & \textbf{49.7} \\
\midrule
\rowcolor{lightgray}
\multicolumn{7}{l}{RxR-CE} \\
VLN\(\circlearrowright\)BERT \cite{hong2022bridging} \textcolor{gray}{\scalebox{0.7}{[CVPR2022]}} & - & - & - & 8.98 & 27.1 & 22.7 \\
ENP \cite{liu2024vision}  \textcolor{gray}{\scalebox{0.7}{[NeurIPS2024]}}  & 5.10 & 62.0  & 51.2 & 5.51 & 55.3 & 45.1 \\
NaVILA \cite{cheng2024navila} \textcolor{gray}{\scalebox{0.7}{[RSS2025]}}  & - & - & - &  6.77 & 49.3 &  44.0 \\
\hline
ETPNav \cite{an2024etpnav} \textcolor{gray}{\scalebox{0.7}{[TPAMI2024]}} & 5.39 & 60.0 & 49.1 & 5.96 & 53.8 & 43.9 \\
ETPNav + VIL (Ours) & \textbf{4.89} & \textbf{63.5} & \textbf{53.0} & \textbf{5.62} & \textbf{55.6} & \textbf{46.2} \\
\bottomrule
\end{tabular}%
}
\end{table}

\textbf{Performance on RxR-CE.} The RxR-CE dataset is much larger than R2R-CE, providing a stronger test of model scalability. As shown in Table \ref{tab:combined_r2r_rxr_ce}, applying VIL under the \textit{Varied Viewpoint} setting yields clear improvements. ETPNav + VIL outperforms the base ETPNav on both val-seen and val-unseen, improving nDTW by 10\%-13\%, OSR by 9\%-13\%, SR by 11\%-15\%, and SPL by 10\%-15\%.

\subsection{Performance under the standard viewpoint}\label{sec:vlnce}

Although VIL is trained with varied viewpoints, it does not degrade performance under the standard VLN-CE setting. As shown in Table~\ref{tab:std-viewpoint}, VIL maintains or slightly improves navigation metrics compared to the base models, demonstrating robustness to viewpoint variations. On R2R-CE val-unseen, ETPNav + VIL increases SR by 1.5 and SPL by 0.8, while on RxR-CE val-unseen, ETPNav + VIL improves SPL by 2.3. These results show that VIL generalizes well to standard viewpoints, confirming that training with varied viewpoints does not compromise, and can even slightly enhance, performance in the original setting. Moreover, the table also includes state-of-the-art map-free methods, where map-free means that no pre-exploration of environments is used. On RxR-CE, our method not only improves over the base model but also outperforms these state-of-the-art map-free methods across all metrics.

\subsection{Ablation study}\label{sec:ablation} 

\begin{table*}[t]
\centering
\caption{
Ablation study on R2R-CE with \textit{Varied Viewpoint} and  \textit{Standard Viewpoint} setting. The best performance for each metric is highlighted in \textbf{bold}, and the second-best is \underline{underlined}. All ablation settings use the same training configurations, including batch size and total training steps. CL: contrastive learning, WPD: waypoint predictor distillation.
}
\resizebox{\textwidth}{!}{%
\begin{tabular}{l ccc  ccc  ccc ccc ccc}
\toprule
\multirow{3}{*}{Method} & \multirow{3}{*}{retrain} & \multirow{3}{*}{CL} & \multirow{3}{*}{WPD} & \multicolumn{6}{c}{\textit{Varied Viewpoint}}  & \multicolumn{6}{c}{\textit{Standard Viewpoint}} \\
\cmidrule(lr){5-10} \cmidrule(lr){11-16} 
& & & & \multicolumn{3}{c}{val-seen} & \multicolumn{3}{c}{val-unseen} & \multicolumn{3}{c}{val-seen} & \multicolumn{3}{c}{val-unseen} \\
\cmidrule(lr){5-7} \cmidrule(lr){8-10} \cmidrule(lr){11-13} \cmidrule(lr){14-16}
& & & & NE$\downarrow$ & SR$\uparrow$ & SPL$\uparrow$ & NE$\downarrow$ & SR$\uparrow$ & SPL$\uparrow$  & NE$\downarrow$ & SR$\uparrow$ & SPL$\uparrow$  & NE$\downarrow$ & SR$\uparrow$ & SPL$\uparrow$  \\
\midrule
\multirow{5}{*}{ETPNav} & $\times$ & $\times$ & $\times$ & 5.16 & 49.4 & 42.0 & 5.58 & 46.8 & 38.4 & 3.97 & 65.8 & 59.2 & 4.78 & 56.7 & \underline{48.9} \\
& $\checkmark$ & $\times$ & $\times$ & 4.52 & 54.6 & 46.0 & 5.21 & 49.8 & 39.1 & 3.62 & 66.1 & 57.6 & 4.66 & 57.9 & 47.7 \\
& $\checkmark$ & $\checkmark$ & $\times$ & 4.55 & 56.4 & 45.3 & 5.09 & 49.7 & 37.2 & 3.72 & 67.0 & 56.8 & \underline{4.64} & \underline{58.2} & 45.6 \\
& $\checkmark$ & $\times$ & $\checkmark$& \underline{4.27} & \underline{59.6} & \underline{51.8} & \underline{5.01} & \underline{52.6} & \underline{42.7} & \textbf{3.51} & \textbf{68.5} & \textbf{60.5} & \textbf{4.63} & 57.6 & 48.7 \\
& $\checkmark$ & $\checkmark$ & $\checkmark$ & \textbf{4.02} & \textbf{63.6} & \textbf{54.9} & \textbf{4.90} & \textbf{54.6} & \textbf{45.5} & \underline{3.71} & \underline{67.6} & \underline{60.4} & 4.69 & \textbf{58.3} & \textbf{49.7}\\
\bottomrule
\end{tabular}
}
\label{tab:ablation}
\end{table*}

We conduct an ablation study in Table~\ref{tab:ablation} to evaluate the contribution of three components: exposing the model to \textit{Varied Viewpoint} data (retrain), contrastive learning (CL), and waypoint predictor distillation (WPD). \textbf{Is standard fine-tuning sufficient?} Retraining on \textit{Varied Viewpoint} improves varied viewpoint SPL slightly (+0.7), but can slightly harm standard viewpoint SPL (-1.2). \textbf{Effect of WPD.} Removing WPD degrades performance substantially, e.g., val-unseen SPL drops by 8.3 (\textit{Varied Viewpoint}) and 4.1 (\textit{Standard Viewpoint}) \textbf{Effect of CL.} Contrastive learning improves val-unseen SPL by 2.8 (\textit{Varied Viewpoint}) and enhances generalization under the standard viewpoint, increasing SR and SPL by 0.7 and 1.0 respectively. These results show that each component contributes to better navigation, particularly in unseen or varied viewpoint settings. We further confirm the same trends on the larger RxR-CE dataset, though we omit the detailed table due to space constraints.

\subsection{Viewpoint Robustness Analysis}\label{sec:robustness}

\begin{table}[t]
\caption{
Standard deviation $\sigma$ for all metrics across 81 viewpoints on R2R-CE val-unseen. 
}
\label{tab:vv_std}
\centering
\resizebox{0.8\columnwidth}{!}{%
\begin{tabular}{lccccc}
\toprule
Model & $\sigma_{\text{NE}}$ & $\sigma_{\text{nDTW}}$ & $\sigma_{\text{OSR}}$ & $\sigma_{\text{SR}}$ & $\sigma_{\text{SPL}}$ \\
\midrule
ETPNav & 0.82 & 7.83 & 9.12 & 10.54 & 10.79 \\
+ VIL & \textbf{0.28} & \textbf{2.43} & \textbf{3.42} & \textbf{3.66} & \textbf{3.59} \\
\bottomrule
\end{tabular}
}
\end{table}
\textbf{Variance across viewpoint changes.} 
We evaluate robustness by sampling 81 fixed viewpoints covering height and angle variations. To quantify this consistency, we compute the standard deviation of each metric across the 81 configurations. As in Table~\ref{tab:vv_std}, our method substantially reduces variance compared to the baseline (e.g., SPL std drops by 65\%), demonstrating
that VIL not only improves average performance, but also stabilizes behavior under spatial perturbation.

\textbf{Evaluation with fixed real-robot camera placements.} 
To further examine viewpoint robustness under realistic configurations, we evaluate navigation performance using camera placements corresponding to three robots: Stretch RE-1, Stretch RE-1 (Factory), and LoCoBot. Only the camera placement is adopted, not the robot hardware, as our inputs are 360$^\circ$ RGB-D images. Both ETPNav and ETPNav+VIL are evaluated on the full val-seen and val-unseen splits.

\begin{table}[t]
\caption{
Navigation performance using fixed camera placements from real robots. Metrics: NE$\downarrow$, SR$\uparrow$, SPL$\uparrow$.
}
\label{tab:realrobot-place}
\centering
\resizebox{\linewidth}{!}{%
\begin{tabular}{l l ccc ccc}
\toprule
\multirow{2}{*}{Robot} & \multirow{2}{*}{Method} & \multicolumn{3}{c}{val-seen} & \multicolumn{3}{c}{val-unseen} \\
\cmidrule(lr){3-5} \cmidrule(lr){6-8} 
 & & NE  & SR & SPL & NE & SR & SPL \\
\midrule
StrRE-1 & ETPNav & 7.50 & 21.9 & 14.5 & 7.08 & 23.1 & 15.5 \\
 & +VIL & 5.37 & 47.8 & 40.1 & 5.83 & 43.7 & 35.4 \\
StrRE-1 (Fac) & ETPNav & 7.38 & 23.8 & 16.1 & 7.09 & 22.7 & 14.8 \\
 & +VIL & 5.57 & 47.9 & 39.6 & 6.00 & 40.3 & 32.0 \\
LoCoBot & ETPNav & 4.38 & 61.3 & 51.6 & 5.17 & 54.2 & 42.8 \\
 & +VIL & 3.80 & 66.7 & 56.4 & 4.91 & 56.0 & 45.7 \\
\bottomrule
\end{tabular}%
}
\end{table}

VIL consistently improves navigation performance across all robot placements. For example, SPL on val-unseen increases by over 20\% for Stretch RE-1, demonstrating that VIL generalizes effectively to diverse camera configurations.

\subsection{Out-of-Distribution Viewpoint Generalization}\label{sec:ood}

In addition to robustness within the training range, we study extrapolation to unseen viewpoints. We compare two training distributions: (i) a \textit{large} range $\mathcal{U}([-0.5\text{m}, 0.5\text{m}]) \times \mathcal{U}([-30^\circ, 30^\circ])$, where the test viewpoints lie on the distribution boundary (thus still in-distribution), and (ii) a \textit{small} range $\mathcal{U}([-0.4\text{m}, 0.4\text{m}]) \times \mathcal{U}([-20^\circ, 20^\circ])$, where the same test viewpoints fall outside the training support (true OOD). We evaluate on two extreme configurations, $(-0.5\text{m}, 30^\circ)$ and $(0.5\text{m}, -30^\circ)$.

\begin{table}[ht]
\caption{Performance under OOD viewpoints on R2R-CE val-seen/unseen.}
\label{tab:ood}
\centering
\resizebox{\linewidth}{!}{%
\begin{tabular}{l l ccc ccc}
\toprule
\multirow{2}{*}{Config} & \multirow{2}{*}{Method} & \multicolumn{3}{c}{val-seen} & \multicolumn{3}{c}{val-unseen} \\
\cmidrule(lr){3-5} \cmidrule(lr){6-8} 
 & & NE  & SR & SPL & NE & SR & SPL \\
\midrule
$(-0.5\text{m}, 30^\circ)$ & ETPNav & 6.51 & 36.2 & 26.2 & 6.30 & 36.3 & 25.9 \\
 & +VIL (large) & 4.69 & 54.8 & 46.4 & 5.18 & 50.2 & 41.0 \\
 & +VIL (small) & 4.78 & 54.0 & 45.6 & 5.43 & 47.5 & 39.2 \\
\midrule
$(0.5\text{m}, -30^\circ)$ & ETPNav & 5.30 & 51.7 & 42.5 & 5.50 & 46.6 & 37.9 \\
 & +VIL (large) & 4.13 & 61.2 & 52.7 & 5.14 & 52.3 & 43.7 \\
 & +VIL (small) & 4.09 & 61.7 & 54.3 & 5.14 & 52.1 & 44.2 \\
\bottomrule
\end{tabular}%
}
\end{table}

Across both OOD configurations, VIL outperforms the baseline by a large margin. Notably, even when trained on the reduced viewpoint range, VIL improves SPL by $+13.3$ on $(-0.5\text{m}, 30^\circ)$ and $+6.3$ on $(0.5\text{m}, -30^\circ)$ (val-unseen). Compared with the large-range model, the small-range model shows only a slight drop in performance, indicating that VIL maintains robustness even with limited viewpoint diversity during training.

\subsection{Real-robot Evaluation}\label{ref:real_robot}

We further validate VIL in real-world settings using a TurtleBot v2 platform. The robot is equipped with a RICOH THETA X 360 RGB camera, a Ouster OS0 Rev6 LiDAR, and an onboard Intel NUC 11 mini-PC (i7-1165G7 CPU, 8 GB RAM). The sensors are extrinsically calibrated following \cite{koide2023general} to align LiDAR and camera frames. This setup yields 12 aligned RGB-D views covering 360°. Unlike prior work such as GVNav \cite{li2025ground}, which relies on rotating a monocular RGB-D camera to synthesize panoramic inputs, our design directly produces 360° RGB-D observations by fusing a panoramic RGB sensor with a 360° LiDAR. 

\begin{figure}[t]
    \centering
    \includegraphics[width=0.6\linewidth]{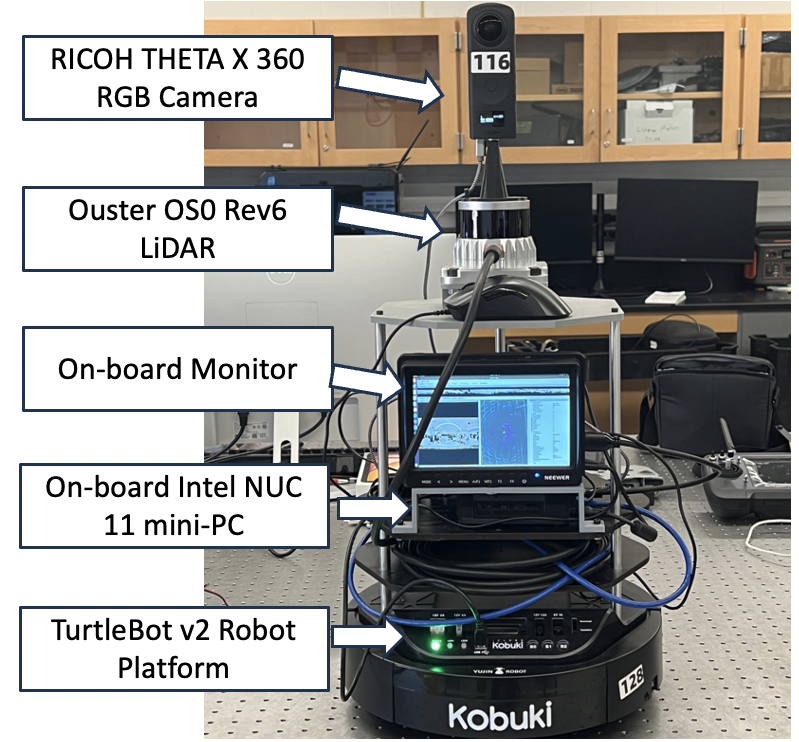}
    \caption{The robot platform used in our experiments.}
    \label{fig:robot_setup}
\end{figure}

The client robot collects RGB-D observations and communicates with a remote server via ROS 2 over a VPN. On the server, our model processes incoming images in real time using an NVIDIA A5000 GPU and outputs navigation actions, which are transmitted back for execution. The real-world experiment is a zero-shot evaluation: the developed model is trained entirely in simulation, using the varied-viewpoint R2R-CE setup. Specifically, the simulation training distribution covered camera heights between 0.75m and 1.75m, while the real robot’s camera was measured at 0.7m, representing an out-of-distribution embodiment.

\begin{figure}[t]
    \centering
    \includegraphics[width=0.8\linewidth]{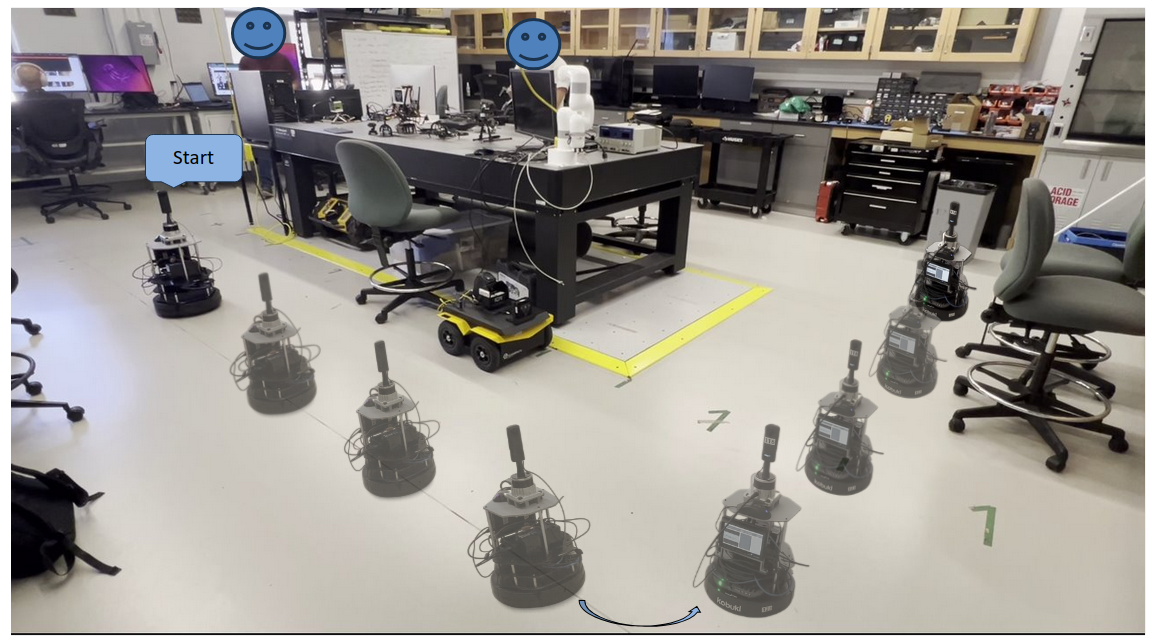}
    \caption{Real world demo of our proposed VIL.}
    \label{fig:real-world-demo}
\end{figure}

We evaluate in two indoor environments: an office and a lounge. Each setting includes 5 instructions, repeated 5 times from different starting locations. Results in Table~\ref{tab:realrobot-res} show that VIL improves navigation robustness across environments. 

\begin{table}[t]
\centering
\caption{Real-robot evaluation in two environments. We report success rate (SR) before and after applying VIL.}
\label{tab:realrobot-res}
\begin{tabular}{lcc}
\toprule
Environment & SR (ETPNav) & SR (+VIL) \\
\midrule
Office  & 28 & \textbf{44} \\
Lounge  & 20 & \textbf{48} \\
\bottomrule
\end{tabular}
\end{table}

These results confirm that VIL consistently enhances the robustness of navigation in real-world deployments, supporting its practicality for embodied agents beyond simulation.

\subsection{Computational Efficiency}\label{sec:eff}

Beyond performance improvements, we examine the training and inference cost of VIL compared to the baseline. 
While ETPNav requires extensive pre-training and fine-tuning stages (around 11.5 days in total), VIL post-training converges in only 48 hours. 
This corresponds to roughly 14\% of the full training time.  The full VIL model has 335.21M total parameters with 143.21M trainable parameters, while the corresponding baseline has 317.31M total parameters with 142.93M trainable parameters. The difference is marginal, confirming that the additional modules introduced by VIL are lightweight.

We also observe that the peak GPU memory usage increases only marginally (from $\sim$6000 MB to 6200--6300 MB with the same batch size). 
At inference, the overhead is negligible since VIL adds only a single linear projection, resulting in no distinguishable difference in per-step runtime. 
These results confirm that VIL is both training-efficient and deployment-friendly, making it practical for real-world navigation.

\section{Conclusion}
We introduced $V^{2}$-VLNCE, a varied-viewpoint scenario to evaluate robustness of VLNCE policies. To address viewpoint sensitivity, we proposed View Invariant Learning (VIL), which improves generalization in both $V^{2}$-VLNCE and standard VLNCE. Real-robot experiments further confirm its effectiveness, showing that VIL is a practical solution for simulated and real-world navigation.

\section*{ACKNOWLEDGMENT}
The authors thank Laura McCrackin (University of Waterloo), Yurun Chen (Shanghai Jiao Tong University and Eastern Institute of Technology, Ningbo), Xinzhu Fu (National University of Singapore), Rui Wang (Hohai University), Jiangran Lyu (Peking University), and Tianyi Hu for helpful discussions.

\bibliographystyle{IEEEtran}
\bibliography{ref}

\end{document}